\documentclass[11pt,letterpaper]{article}
\usepackage{hyperref}
\usepackage{setspace}
\usepackage{naaclhlt2016}
\usepackage{times}
\usepackage{latexsym}

\usepackage{booktabs}
\usepackage{siunitx}
\usepackage{enumitem}

\usepackage{amsmath}
\usepackage{amssymb}
\usepackage{amsthm}
\usepackage{stmaryrd}

\usepackage{tikz}
\usepackage{mdframed}
\usepackage{relsize}
\usepackage{tabu}
\usepackage{xcolor}
\definecolor{silver}{RGB}{214,214,214}
\usepackage{verbatimbox}
\usepackage{xspace}

\newcommand{\grad}{\nabla}
\newcommand{\expect}{\mathbb{E}}

\newcommand{\lparam}{\theta_\ell}

\newcommand{\mrl}[1]{{\texttt{\smaller #1}}}
\newcommand{\denote}[1]{\llbracket #1 \rrbracket}

\newcommand{\atttype}{\underline{Attention}\xspace}
\newcommand{\labtype}{\underline{Labels}\xspace}

\definecolor{answergreen}{RGB}{64,128,0}
\definecolor{answerred}{RGB}{128,0,0}
\newcommand{\correct}[1]{\textcolor{answergreen}{\textbf{#1}}}
\newcommand{\incorrect}[1]{\textcolor{answerred}{\textbf{#1}}}

\title{Learning to Compose Neural Networks for Question Answering}

\taburulecolor{silver}
\author{
  Jacob Andreas \and Marcus Rohrbach \and Trevor Darrell \and Dan Klein \\
  Department of Electrical Engineering and Computer Sciences \\
  University of California, Berkeley \\
  {\tt \{jda,rohrbach,trevor,klein\}@eecs.berkeley.edu}
}

\date{}

\setlist{noitemsep}

\naaclfinalcopy

\begin{document}
\maketitle

\begin{abstract}
  We describe a question answering model that applies to both images and
  structured knowledge bases.
  The model uses natural language strings to
  automatically assemble neural networks from a collection of
  composable modules. Parameters for these modules are learned
  jointly with network-assembly parameters via reinforcement 
  learning, with only (world, question, answer) triples as supervision. Our 
  approach, which we term a \emph{dynamic neural module network}, 
  achieves state-of-the-art results on benchmark datasets in both 
  visual and structured domains.
\end{abstract}

\section{Introduction}
\label{sec:intro}

This paper presents a compositional, attentional model for answering questions
about a variety of world representations, including images and structured
knowledge bases. The model translates from questions to dynamically assembled
neural networks, then applies these networks to world representations (images or
knowledge bases) to produce answers. We take advantage of two largely independent
lines of work: on  one hand, an extensive literature on answering questions by 
mapping from strings to logical representations  of meaning; on the other, 
a series of recent successes in deep neural models for image recognition and 
captioning. By constructing neural networks instead of logical forms, our
model leverages the best aspects of both linguistic compositionality and continuous 
representations.

\begin{figure}
  \includegraphics[
    width=\columnwidth,
    trim=0 5.5cm 5.5cm 0cm, 
    clip
  ]{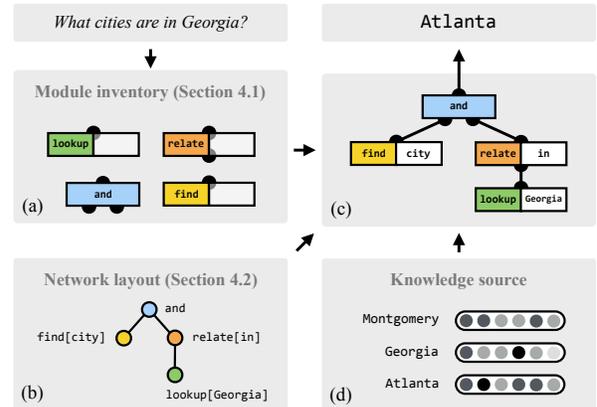}
  \caption{A learned syntactic analysis (a) is used to assemble a 	
  collection of neural modules (b) into a deep neural network (c), 
  and applied to a world representation (d) to produce an answer.}
  \label{fig:teaser}
  \vspace{-1mm}
\end{figure}

Our model has two components, trained jointly: first, a collection of neural
``modules'' that can be freely composed (\autoref{fig:teaser}a); second, a network layout predictor
that assembles modules into complete deep networks tailored to each question (\autoref{fig:teaser}b).
Previous work has used manually-specified modular structures for visual learning 
\cite{Andreas15NMN}. Here we:
\begin{itemize}
	\item \emph{learn} a network structure predictor jointly with module 
    	  parameters themselves \\    \item \emph{extend} visual primitives from previous work to reason over
    	  structured world representations
\end{itemize}
Training data consists of (world, question, answer) triples: our approach requires no 
supervision of network layouts.
We achieve state-of-the-art performance on two markedly different question 
answering tasks: one with questions about natural images, and another with more compositional questions about United States geography.\footnote{We have released our code at 
  {\smaller\url{http://github.com/jacobandreas/nmn2}}}

\section{Deep networks as functional programs}
\label{sec:programs}

We begin with a high-level discussion of the kinds of composed networks we would like to learn. 
\pagebreak

\begin{figure}
  \centering

  \includegraphics[
  	width=.9\columnwidth,
    trim=0.5cm 7.3cm 11.5cm 0.5cm,
    clip
  ]{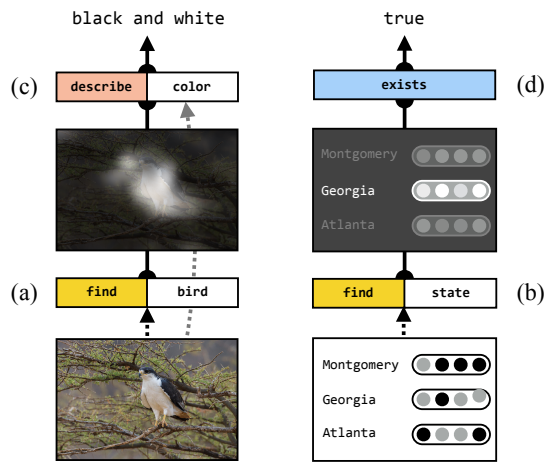}
  \caption{Simple neural module networks, corresponding to the questions
  \emph{What color is the bird?}\ and \emph{Are there any states?} 
  (a)~A neural \mrl{find} module for
  computing an attention over pixels. (b) The same operation applied to a
  knowledge base. (c)~Using an attention produced by a lower module to identify
  the color of the region of the image attended to. (d) Performing quantification
  by evaluating an attention directly.}
  \label{fig:examples}
  \end{figure}

\newcite{Andreas15NMN} describe a heuristic approach for
decomposing visual question answering tasks into sequence of modular sub-problems. For example,
the question \emph{What color is the bird?}\ might be answered in two steps: first, ``where 
is the bird?'' (\autoref{fig:examples}a), second, ``what color is that part of the image?'' 
(\autoref{fig:examples}c).
This first step, a generic \textbf{module} called \mrl{find}, can be expressed as a fragment of a neural 
network that maps from image features and a lexical item (here \emph{bird}) to a distribution over pixels. 
This operation is commonly referred to as the \emph{attention mechanism}, and is a standard tool for 
manipulating images \cite{Xu15SAT} and text representations \cite{Hermann15AttQA}.

The first contribution of this paper is an extension and generalization of this mechanism 
to enable fully-differentiable reasoning 
about more structured semantic representations. 
\autoref{fig:examples}b shows how the same module can be used to focus on the entity \emph{Georgia}
in a non-visual grounding domain; more generally, by representing every entity in the
universe of discourse as a feature vector, we can obtain a distribution 
over entities that corresponds roughly to a logical set-valued denotation.

Having obtained such a distribution, existing neural approaches use it to 
immediately compute a weighted average of image features and project back into a labeling decision---a 
\mrl{describe} module
(\autoref{fig:examples}c). But the logical perspective suggests a number of novel modules that might
operate on attentions: e.g.\ combining them (by analogy to conjunction or disjunction) or inspecting 
them directly without a return to feature space (by analogy to quantification, 
\autoref{fig:examples}d). These modules are discussed in detail in \autoref{sec:model}. 
Unlike their formal counterparts, they are differentiable end-to-end, facilitating their 
integration into learned models. Building on previous work, we learn behavior
for a collection of heterogeneous modules from (world, question, answer) triples.

The second contribution of this paper is a model for learning to assemble such modules
compositionally. Isolated modules are of limited use---to obtain expressive power comparable
to either formal approaches or monolithic deep networks, they must be composed into larger 
structures. \autoref{fig:examples} shows simple examples of composed structures, but for 
realistic question-answering tasks, even larger networks are required.
Thus our goal is to automatically induce variable-free, tree-structured computation descriptors. 
We can use a familiar functional notation from formal semantics (e.g. Liang et al., 2011) to 
represent these computations.\footnote{But note that unlike formal semantics, the behavior of the primitive 
functions here is itself unknown.} We write the two examples in \autoref{fig:examples} as

{\smaller
\begin{verbatim}
    (describe[color] find[bird])
\end{verbatim}
}
\noindent and
{\smaller 
\begin{verbatim}
    (exists find[state])
\end{verbatim}
}

\noindent respectively. These are \textbf{network layouts}: they specify a structure for arranging modules (and their lexical parameters) into a complete network. \newcite{Andreas15NMN} use hand-written 
rules to deterministically  transform dependency trees into layouts, and are restricted to producing
simple structures like the above for non-synthetic data. For full generality, 
we will need to solve harder problems, like transforming \emph{What cities are in Georgia?}
(\autoref{fig:teaser}) into

{\smaller 
\begin{verbatim}
    (and
        find[city]
        (relate[in] lookup[Georgia]))
\end{verbatim}
}
\noindent In this paper, we present a model for learning to select such structures from a set of automatically
generated candidates. We call this model a \textbf{dynamic neural module network}.

\section{Related work}
\label{sec:related}

There is an extensive literature on database question answering, in which
strings are mapped to logical forms, then evaluated by a black-box execution 
model to produce answers. Supervision may be provided either by annotated logical forms
\cite{Wong07WASP,Kwiatkowski10UBL,Andreas13SPMT} or from (world,
question, answer) triples alone \cite{Liang11DCS,Pasupat15Tables}. In general the
set of primitive functions from which these logical forms can be assembled is
fixed, but one recent line of work focuses on inducing new predicates functions
automatically, either from perceptual features \cite{Krish2013Grounded} or 
the underlying schema \cite{Kwiatkowski13Ontology}. The model we describe in this paper
has a unified framework for handling both the perceptual and schema cases, and
differs from existing work primarily in learning a differentiable
execution model with continuous evaluation results.

Neural models for question answering are also a subject of current interest.
These include approaches that model the task directly as a multiclass classification 
problem  \cite{Iyyer14Factoid}, models that attempt to embed questions and answers
in a shared vector space \cite{Bordes14GraphEmbedding} and attentional models that select words from
documents sources \cite{Hermann15AttQA}. Such approaches generally require
that answers can be retrieved directly based on surface linguistic features, 
without requiring intermediate computation. A more structured approach
described by \newcite{Yin15NeuralTable} learns a query execution model
for database tables without any natural language component. Previous efforts
toward unifying formal logic and representation learning include those of 
\newcite{Grefenstette13Logic}, \newcite{Krishnamurthy13CompVector},
\newcite{Lewis13DistributionalLogical}, and \newcite{Beltagy13Markov}.

The visually-grounded component of this work relies on recent advances
in convolutional networks for computer vision
\cite{Simonyan14VGG}, and in particular the fact that late
convolutional layers in networks trained for image recognition contain rich features 
useful for other vision tasks while preserving spatial information.
These features have been used for both image captioning \cite{Xu15SAT} and
visual QA \cite{Yang15AttVQA}.

Most previous approaches to visual question answering either apply a recurrent
model to deep representations of both the image and the question
\cite{Ren15VQA,Malinowski15VQA}, or use the question to compute an attention
over the input image, and then answer based on both the question and the image
features attended to \cite{Yang15AttVQA,Xu15AttVQA}.  Other approaches include
the simple classification model described by \newcite{Zhou15ClassVQA} and the
dynamic parameter prediction network described by \newcite{Noh15DPPVQA}.  All of
these models assume that a fixed computation can be performed on the image and
question to compute the answer, rather than adapting the structure of the
computation to the question. 

As noted, \newcite{Andreas15NMN} previously considered a simple generalization
of these attentional approaches in  which small variations in the network
structure per-question were permitted, with the structure chosen by
(deterministic) syntactic processing of questions. Other approaches in this
general family include the ``universal parser'' sketched by
\newcite{Bottou14Reasoning}, the graph transformer networks of
\newcite{Bottou97GraphTransformers}, the knowledge-based neural networks of
\newcite{Towell94KBNN} and the recursive neural networks of
\newcite{Socher13CVG}, which use a fixed tree structure to perform further
linguistic analysis without any external world representation.  We are unaware
of previous work that simultaneously learns both parameters
for and structures of instance-specific networks.

\section{Model}
\label{sec:model}

Recall that our goal is to map from questions and world representations to answers. 
This process involves the following variables:
\begin{enumerate}
  \item $w$ a world representation
  \item $x$ a question
  \item $y$ an answer
  \item $z$ a network layout
  \item $\theta$ a collection of model parameters
\end{enumerate}
Our model is built around two distributions:
a \textbf{layout model} $p(z|x;\lparam)$ which chooses a layout for a sentence,
and a \textbf{execution model} $p_z(y|w;\theta_e)$ which applies the network 
specified by $z$ to $w$.

For ease of
presentation, we introduce these models in reverse order. We first imagine that $z$ is always observed, and in \autoref{sec:model:modules} describe how to 
evaluate and learn modules parameterized by $\theta_e$ within fixed structures. In \autoref{sec:model:assemblingNetworks}, we move to the real scenario, where $z$ is unknown. We describe how to predict layouts 
from questions and learn $\theta_e$ and $\lparam$ jointly without layout supervision.

\subsection{Evaluating modules}
\label{sec:model:modules}

Given a layout $z$, we assemble the corresponding modules into a full neural network
(\autoref{fig:teaser}c), and apply it to the knowledge representation. 
Intermediate results flow between modules until an answer is
produced at the root. We denote the output of the network with layout $z$ on
input world $w$ as $\denote{z}_w$; when explicitly referencing the substructure
of $z$, we can alternatively write $\denote{m(h^1, h^2)}$ for a top-level module $m$
with submodule outputs $h^1$ and $h^2$.
We then define the execution model:
\vspace{-1mm}
\begin{equation}
  p_z(y|w) = (\denote{z}_w)_y
  \label{eq:simple-execution}
\end{equation}
\vspace{-1mm}
(This assumes that the root module of $z$ produces a distribution over labels $y$.)
The set of possible layouts $z$ is restricted by module \emph{type constraints}:
some modules (like \mrl{find} above) operate directly on the input representation,
while others (like \mrl{describe} above) also depend on input from specific earlier 
modules. Two base types are considered in this paper are \atttype (a distribution over pixels 
or entities) and \labtype (a distribution over answers).

Parameters are tied across multiple instances of the same module, so different 
instantiated networks may share some parameters but not others.
Modules have both \emph{parameter arguments} (shown in square brackets) and ordinary 
inputs (shown in parentheses). Parameter arguments, like the running \mrl{bird} example
in \autoref{sec:programs}, are provided by the layout, and
are used to specialize module behavior for particular lexical items. Ordinary inputs
are the result of computation lower in the network. In addition to parameter-specific 
weights, modules have global weights shared across all instances of the module (but not
shared with other modules). We write $A, a, B, b, \dots$ for global weights and 
$u^i, v^i$ for weights associated with the parameter argument $i$. $\oplus$ and $\odot$ denote
(possibly broadcasted) elementwise addition and multiplication respectively. The complete set of
global weights and parameter-specific weights constitutes $\theta_e$.
\emph{Every} module has access to the world representation, represented as a 
collection of vectors $w^1, w^2, \dots$ (or $W$ expressed as a matrix). The nonlinearity $\sigma$ denotes a rectified 
linear unit.

The modules used in this paper are shown below, with names
and type constraints in the first row and a description of the module's computation following.

{\small
\setlength{\belowdisplayskip}{5pt} \setlength{\belowdisplayshortskip}{5pt}
\setlength{\abovedisplayskip}{5pt} \setlength{\abovedisplayshortskip}{5pt}
\tabulinesep=3mm
\noindent
\begin{tabu}{|p{0.95\columnwidth}|}
	\hline
    
    \textbf{Lookup} \hfill ($\to$ \atttype) \newline
	\mrl{lookup[$i$]} produces an attention focused entirely at the index $f(i)$,
	where the relationship $f$ between words and positions in the input map is 
    known ahead of time (e.g. string matches on database fields).
	\begin{equation}
		\denote{\mrl{lookup[$i$]}} = e_{f(i)}
	\end{equation}
	where $e_i$ is the basis vector that is $1$ in the $i$th position and 0 elsewhere. \\
    
    \hline    
    \setlength{\belowdisplayskip}{-5pt}
	\textbf{Find} \hfill ($\to$ \atttype) \newline
	\mrl{find[$i$]} computes a distribution over indices by concatenating the parameter 
	argument with each  position of the input feature map, and passing the concatenated 
	vector through a MLP:
	\begin{equation}
		\denote{\mrl{find[$i$]}} = \textrm{softmax}(a \odot \sigma(B v^i \oplus C W \oplus d))
	\end{equation} \\
    
    \hline    
    \setlength{\belowdisplayskip}{-5pt}
	\textbf{Relate} \hfill (\atttype $\to$ \atttype) \newline
	\mrl{relate} directs focus from one region of the input to another. It behaves much
    like the \mrl{find} module, but also conditions its behavior on the current region
    of attention $h$. Let 
         $\bar{w}(h) = \sum_k h_k w^k$, where $h_k$ is the $k^{th}$ element of $h$. Then,
         {\begin{align}
            & \denote{\mrl{relate[$i$]}(h)} = \textrm{softmax}(a \ \odot \nonumber \\ 
            &\qquad\sigma(B v^i \oplus C W \oplus D\bar{w}(h) \oplus e))
         \end{align}}
         \\
    \hline
        
	\textbf{And} \hfill (\atttype{}* $\to$ \atttype) \newline
    \setlength{\belowdisplayskip}{0pt}
    \mrl{and} performs an operation analogous to set intersection
    for attentions. The analogy to probabilistic logic suggests multiplying probabilities:
    \begin{equation}
    \denote{\mrl{and}(h^1, h^2, \ldots)} = h^1 \odot h^2 \odot \cdots
    \end{equation}\\
    
    \hline    
	\textbf{Describe} \hfill (\atttype $\to$ \labtype) \newline
    \setlength{\belowdisplayskip}{-5pt}
	\mrl{describe[$i$]} computes a weighted average of $w$ under the input attention. This average is then used to predict an answer representation. With $\bar{w}$ as above,
	\begin{equation}
	    \denote{\mrl{describe[$i$]}(h)} = \textrm{softmax}(A \sigma(B \bar{w}(h) + v^i))
	\end{equation} \\
    
    \hline
    
	\textbf{Exists} \hfill (\atttype $\to$ \labtype) \newline
    \setlength{\belowdisplayskip}{-5pt}
	    \mrl{exists} is the existential quantifier, and inspects the incoming
	attention directly to produce a label, rather than an intermediate feature
	vector like \mrl{describe}:
	\begin{equation}
		\denote{\mrl{exists]}(h)} = \textrm{softmax}\Big(\big(\max_k h_k\big)a  + b\Big)
	\end{equation} \\
    \hline
\end{tabu}
}

\pagebreak

With $z$ observed, the model we have described so far corresponds largely
to that of \newcite{Andreas15NMN}, though the module inventory is different---in
particular, our new \mrl{exists} and \mrl{relate} modules do not depend on the
two-dimensional spatial structure of the input. This enables generalization to non-visual 
world representations.

Learning in this simplified setting is straightforward. Assuming the top-level 
module in each layout is a \mrl{describe} or \mrl{exists} module, the fully-
instantiated network corresponds to a distribution over labels conditioned on
layouts. To train, we maximize
    \\    $
    \sum_{(w,y,z)} \log p_z(y|w;\theta_e)
$
directly.
This can  be understood as a 
parameter-tying scheme, where the decisions about which 
parameters to tie are governed by the observed layouts $z$.

\subsection{Assembling networks}
\label{sec:model:assemblingNetworks}
\begin{figure}
  \centering
    \includegraphics[
    width=.8\columnwidth,
    trim=0.5cm 0.5cm 11.5cm 0.5cm,
    clip
  ]{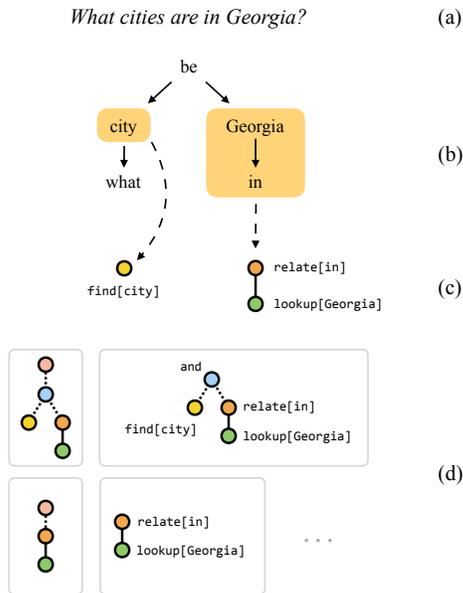}
  \caption{Generation of layout candidates. The input sentence (a) is represented
  as a dependency parse (b). Fragments of this dependency parse are then 	
  associated with appropriate modules (c), and these fragments are assembled 
  into full layouts (d).}
  \label{fig:layout}
  \end{figure}

Next we describe the layout model $p(z|x;\lparam)$.
We first use a fixed syntactic parse to generate a small set of candidate
layouts, analogously to the way a semantic grammar 
generates candidate semantic parses in previous work \cite{Berant14Paraphrasing}. 

A semantic parse differs from a syntactic parse in two primary ways.
First, lexical items must be mapped onto a (possibly smaller) set of 
semantic primitives. Second, these semantic primitives must be combined into a 
structure that closely, but not exactly, parallels the structure provided by syntax.
For example, \emph{state} and \emph{province} might need to be identified with the
same field in a database schema, while \emph{all states have a capital} might need
to be identified with the correct (\emph{in situ}) quantifier scope. 

While we cannot avoid the structure selection problem, continuous representations
simplify the lexical selection problem. For modules that accept a vector parameter,
we associate these parameters with \emph{words} rather than semantic tokens,
 and thus turn the combinatorial optimization problem associated with lexicon induction 
into a continuous one. Now, in order to learn that \emph{province} and \emph{state} 
have the same denotation, it is sufficient to learn that their associated parameters 
are close in some embedding space---a task amenable to gradient descent.
(Note that this is easy only in an optimizability sense, and not an information-theoretic
one---we must still learn to associate each independent lexical item with the correct
vector.) The remaining combinatorial problem is to arrange the provided lexical items 
into the right computational structure. In this respect, layout prediction is 
more like syntactic parsing  than ordinary semantic parsing, and we can rely on an 
off-the-shelf syntactic parser to get most of the way there. In this work, syntactic
structure is provided by the Stanford dependency parser \cite{DeMarneffe08Deps}.

The construction of layout candidates is depicted in \autoref{fig:layout}, and proceeds as follows:
\begin{enumerate}
	\item Represent the input sentence as a dependency tree.
    \item Collect all nouns, verbs, and prepositional phrases that are attached directly to a wh-word or copula.
    \item Associate each of these with a layout fragment: Ordinary nouns and verbs are mapped to a single \mrl{find} module.
    	  Proper nouns to a single \mrl{lookup} module. Prepositional phrases are mapped to a depth-2 fragment, with a \mrl{relate} 
          module for the preposition above a \mrl{find} module for the enclosed head noun. 
    \item Form subsets of this set of layout fragments. For each subset, construct a layout candidate by joining all 
    	  fragments with an \mrl{and} module, and inserting either a \mrl{measure} or \mrl{describe} module at the top
          (each subset thus results in two parse candidates.) 
\end{enumerate}

All layouts resulting from this process feature a relatively flat tree
structure with at most one conjunction and one quantifier.
This is a strong simplifying assumption, but appears sufficient to cover
most of the examples that 
appear in both of our tasks. As our approach includes both categories, relations
and simple quantification, the range of phenomena considered is generally broader than
previous perceptually-grounded QA work \cite{Krish2013Grounded,Matuszek12Grounded}.

Having generated a set of candidate parses, we need to score them. This is a
ranking problem; as in the rest of our approach, we solve it using standard
neural machinery. In particular, we produce an LSTM representation of the question, a 
feature-based representation of the query, and pass both representations through a multilayer perceptron (MLP). 
The query feature vector includes indicators on the number of modules of each type present, as well as their associated parameter arguments.
While one can easily imagine a more sophisticated parse-scoring model, this simple approach works well for our tasks.

Formally, for a question $x$, let $h_q(x)$ be an LSTM encoding of the question 
(i.e. the last hidden layer of an LSTM applied word-by-word to the input question). 
Let $\{z_1, z_2, \ldots\}$ be the proposed layouts for $x$, and let
$f(z_i)$ be a feature vector representing the $i$th layout. Then the score
$s(z_i|x)$ for the layout $z_i$ is
\begin{equation}
	s(z_i|x) = a^\top \sigma(B h_q(x) + C f(z_i) + d)
    \label{eq:layout-score}
\end{equation}
i.e.\ the output of an MLP with inputs $h_q(x)$ and $f(z_i)$, and parameters
$\lparam = \{a, B, C, d\}$. Finally, we normalize these scores to obtain a
distribution:
\begin{equation}
	p(z_i|x;\lparam) = e^{s(z_i|x)} \Big/ \sum_{j=1}^n e^{s(z_j|x)}
\end{equation}

Having defined a layout selection module $p(z|x;\lparam)$ and a network execution
model $p_z(y|w;\theta_e)$, we are ready to define a model for predicting answers
given only (world, question) pairs. The key constraint is that we want to minimize
evaluations of $p_z(y|w;\theta_e)$ (which involves expensive application of a deep network
to a large input representation), but can tractably evaluate $p(z|x;\lparam)$ for all
$z$ (which involves application of a shallow network to a relatively small set of
candidates). This is the opposite of the situation usually encountered semantic
parsing, where calls to the query execution model are fast but the set of candidate
parses is too large to score exhaustively. 

In fact, the problem more closely resembles
the scenario faced by agents in the reinforcement learning setting (where it is cheap 
to score actions, but potentially expensive to execute them and obtain rewards). 
We adopt a common approach from that literature, and
express our model as a stochastic policy. Under this policy, we first \emph{sample} a 
layout $z$  from a distribution $p(z|x;\lparam)$, and then apply $z$ to the knowledge 
source and obtain a distribution over answers $p(y|z,w;\theta_e)$.

After $z$ is chosen, we can train the execution model directly by maximizing $\log
p(y|z,w;\theta_e)$ with respect to $\theta_e$ as before (this is ordinary
backpropagation).  Because the hard selection of $z$ is non-differentiable, we
optimize $p(z|x;\lparam)$ using a policy gradient method.  The gradient of the reward surface $J$ with 
respect to the  parameters of the policy is
\begin{equation}
  \grad J(\lparam) = \expect[ \grad \log p(z|x;\lparam) \cdot r ] 
\end{equation}
(this is the {\sc reinforce} rule \cite{Williams92Reinforce}).
Here the expectation is taken with
respect to rollouts of the policy, and $r$ is the reward. Because our goal is to
select the network that makes the most accurate predictions, we take the
reward to be identically the negative log-probability from the execution phase,
i.e.
\begin{equation}
  \expect [(\grad \log p(z|x;\lparam)) \cdot \log p(y|z,w;\theta_e)]
\end{equation}
Thus the update to the layout-scoring model at each timestep is simply the
gradient of the log-probability of the chosen layout, scaled by the accuracy of
that layout's predictions.
At training time, we approximate the expectation with a single rollout, so at
each step we update $\lparam$ in the direction
$
  (\grad \log p(z|x;\lparam)) \cdot \log p(y|z,w;\theta_e)
$
for a single $z \sim p(z|x;\lparam)$. $\theta_e$ and $\lparam$ are optimized using 
{\sc adadelta} \cite{Zeiler12Adadelta}
with $\rho=0.95,$ $\varepsilon=1\mathrm{e}{-6}$ and gradient clipping at a norm of $10$.

\section{Experiments}
\label{sec:experiments}

The framework described in this paper is general, and we are interested in how well it
performs on datasets of varying domain, size and linguistic complexity. To that end, we
evaluate our model on tasks at opposite extremes of both these criteria: a large 
visual question answering dataset, and a small collection of more structured geography 
questions.

\subsection{Questions about images}

\begin{figure}
  \centering
  \scalebox{0.78}{
    \smaller
    \tabulinesep=3mm
    \begin{tabu}{|p{1.5in}|p{1.0in}|p{0.93in}|}
    	\hline
        \includegraphics[height=1in]{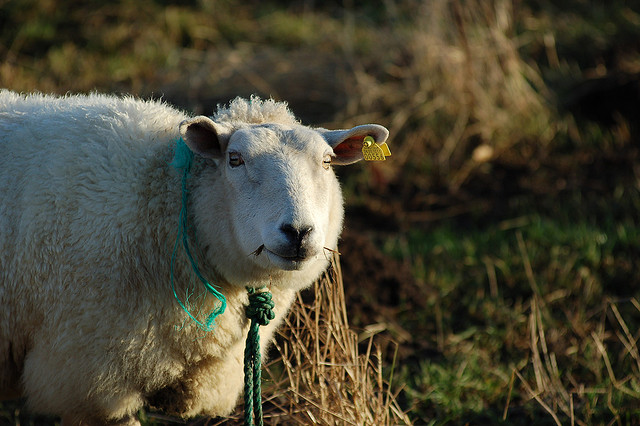} &
        \centering
        \includegraphics[height=1in]{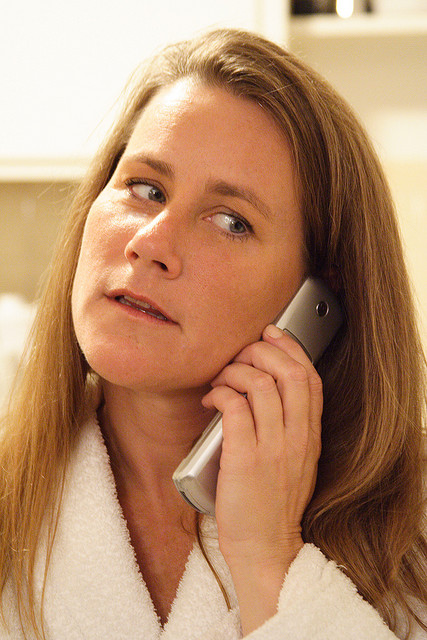} &
        \centering
        \includegraphics[height=1in]{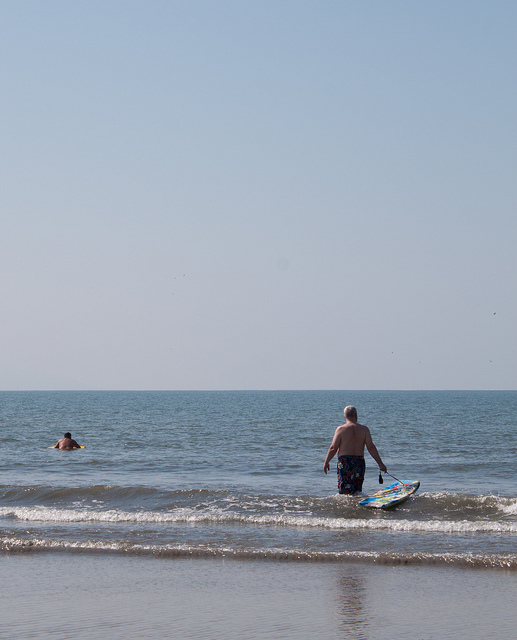}
        \\[-4mm]
        \includegraphics[height=1in]{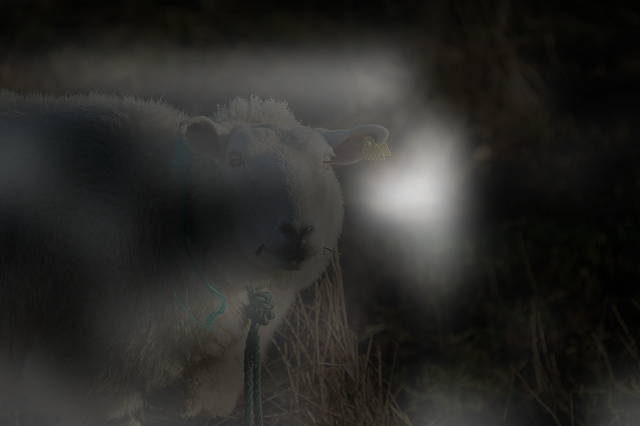} &
        \centering
        \includegraphics[height=1in]{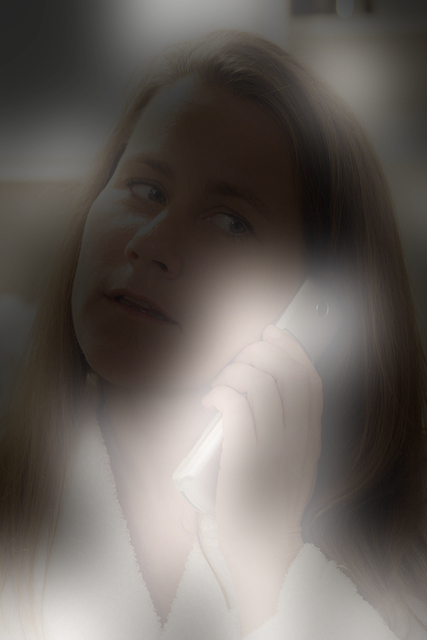} &
        \centering
        \includegraphics[height=1in]{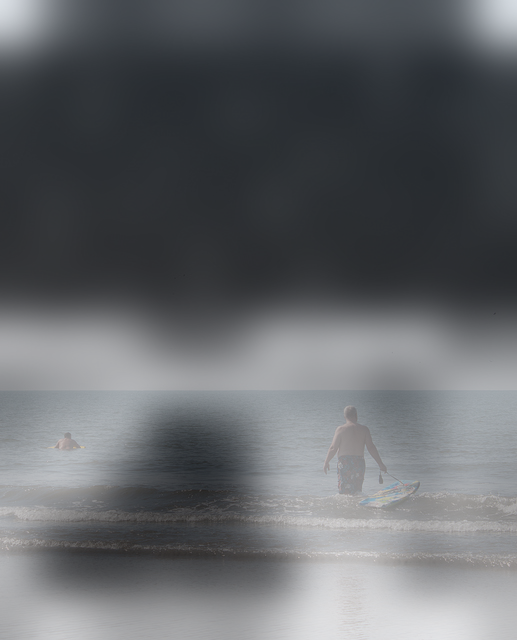}
        \\[-3mm]
        \emph{What is in the sheep's ear?} &
        \raggedright
        \emph{What color is she wearing?} &
        \raggedright
        \emph{What is the man dragging?}
        \\[-3mm]
        \mrl{(describe[what] \newline
        \strut~~~~(and find[sheep] \newline
        \strut~~~~~~~~~find[ear]))} &
        \mrl{(describe[color] \newline
        \strut~~~~find[wear])} &
        \mrl{(describe[what] \newline
        \strut~~~~find[man])}
        \\[-3mm]
        \correct{tag} &
        \correct{white} &
        \incorrect{boat} (board) \\
        \hline
    \end{tabu}
  }
  \vspace{.5mm}
  \caption{Sample outputs for the visual question answering task. The second
  row shows the final attention provided as input to the top-level \mrl{describe} 
  module. For the first two examples, the model produces reasonable parses,
  attends to the correct region of the images (the ear and the woman's 
  clothing), and generates the correct answer. In the third image, the
  verb is discarded and a wrong answer is produced.}
  \label{fig:vqa:qualitative-results}
  \end{figure}

Our first task is the recently-introduced Visual Question Answering challenge (VQA) \cite{Antol15VQA}.
The VQA dataset consists of more than 200,000 images paired with human-annotated questions
and answers, as in \autoref{fig:vqa:qualitative-results}.

We use the VQA 1.0 release, employing the development set for
model selection and hyperparameter tuning, and reporting final results from the
evaluation server on the test-standard set. For the experiments described in this
section, the input feature representations $w_i$ are computed by the the fifth
convolutional layer of a 16-layer VGG\-Net after pooling \cite{Simonyan14VGG}. 
Input images are scaled to 448$\times$448 before computing their representations.
We found that performance on this task was best if the candidate layouts were
relatively simple: only \mrl{describe}, \mrl{and} and \mrl{find} modules
are used, and layouts contain at most two conjuncts.

One weakness of this basic framework is a difficulty modeling prior knowledge about answers
(of the form \emph{most bears are brown}). This kinds of linguistic ``prior'' is essential for the
VQA task, and easily incorporated. We simply introduce an extra hidden layer for recombining the
final module network output with the input sentence representation $h_q(x)$ 
(see \autoref{eq:layout-score}), replacing \autoref{eq:simple-execution} with:
\begin{equation}
  \log p_z(y|w,x) = (A h_q(x) + B \denote{z}_w)_y
\end{equation}
(Now modules with output type \labtype should be understood as producing an answer
embedding rather than a distribution over answers.) This allows the question to influence
the answer directly.

\begin{table}
  \footnotesize
  \centering
  \sisetup{
    table-figures-decimal = 1
  }
    \begin{tabular}{l@{\ }ccccc}
    \toprule
    & \multicolumn{4}{c}{test-dev} & test-std \\
    \cmidrule(lr){2-5} \cmidrule(lr){6-6}
    & Yes/No & Number & Other & All & All \\
    \midrule
    Zhou (2015) & 76.6 & 35.0 & 42.6 & 55.7 & 55.9 \\
    Noh (2015)  & 80.7 & 37.2 & 41.7 & 57.2 & 57.4 \\
    Yang (2015) & 79.3 & 36.6 & 46.1 & 58.7 & 58.9 \\
    NMN         & 81.2 & 38.0 & 44.0 & 58.6 & 58.7 \\
    D-NMN       & 81.1 & 38.6 & 45.5 & 59.4 & \bf 59.4 \\
    \bottomrule
  \end{tabular}
  \vspace{.5mm}
  \caption{Results on the VQA test server. NMN is the parameter-tying
  model from Andreas et al.\ (2015), and D-NMN is the model described
  in this paper.}
    \label{tbl:vqa:quantitative-results}
    \end{table}

Results are shown in \autoref{tbl:vqa:quantitative-results}. 
The use of dynamic networks provides a small gain, most noticeably on "other" questions.
We achieve state-of-the-art results on this task,
outperforming a highly effective visual bag-of-words model \cite{Zhou15ClassVQA},
a model with dynamic network parameter prediction (but fixed network
structure) \cite{Noh15DPPVQA}, a more conventional attentional model \cite{Yang15AttVQA},
and a previous approach using neural module
networks with no structure prediction \cite{Andreas15NMN}. 

Some examples are shown in \autoref{fig:vqa:qualitative-results}. In general, 
the model learns to focus on the correct region of the image, and tends to consider 
a broad window around the region. This facilitates answering questions like 
\emph{Where is the cat?}, which requires knowledge of the surroundings as well as the 
object in question.

\subsection{Questions about geography}

The next set of experiments we consider focuses on GeoQA, a geographical
question-answering task first introduced by \newcite{Krish2013Grounded}.
This task was originally paired with a visual question answering task
much simpler than the one just discussed, and is appealing for a
number of reasons. In contrast to the VQA dataset, GeoQA is quite small,
containing only 263 examples. Two baselines are available: one using a 
classical semantic parser backed by a database, and another which 
induces logical predicates using linear 
classifiers over both spatial and distributional features. This allows
us to evaluate the quality of our model relative to other perceptually
grounded logical semantics, as well as strictly logical approaches.

\begin{table}
  \footnotesize
  \centering
  \sisetup{
    table-figures-decimal = 1
  }
  \begin{tabular}{lcc}
    \toprule
    & \multicolumn{2}{c}{Accuracy} \\
    \cmidrule(){2-3}
    Model & GeoQA & GeoQA+Q \\
    \midrule
    LSP-F & 48\phantom{.0} & -- \\
    LSP-W & 51\phantom{.0} & -- \\
    NMN   & 51.7 & 35.7\\
    D-NMN   & \bf 54.3 & \bf 42.9 \\
    \bottomrule
  \end{tabular}
  \vspace{.5em}
  \caption{Results on the GeoQA dataset, and the GeoQA dataset
  with quantification. Our approach outperforms both a purely
  logical model (LSP-F) and a model with learned perceptual
  predicates (LSP-W) on the original dataset, and a fixed-structure 
  NMN under both evaluation conditions.}
  \label{tbl:geo:quantitative}
  \end{table}

The GeoQA domain consists of a set of entities (e.g.\ states, cities, 
parks) which participate in various relations (e.g.\ north-of, 
capital-of). Here we take the world representation to consist of two pieces:
a set of category features (used by the \mrl{find} module) and a different
set of relational features (used by the \mrl{relate} module). For our experiments,
we use a subset of the features originally used by Krishnamurthy et al.
The original dataset includes no quantifiers, and treats the questions
\emph{What cities are in Texas?}\ and \emph{Are there any cities in Texas?}\
identically. Because we are interested in testing the parser's ability to
predict a variety of different structures, we introduce a new version
of the dataset, GeoQA+Q, which distinguishes these two cases, and expects a
Boolean answer to questions of the second kind.

Results are shown in \autoref{tbl:geo:quantitative}. As in the original work, we report the results
of leave-one-environment-out cross-validation on the set of 10 environments.
Our dynamic model (D-NMN) outperforms both the logical (LSP-F) and perceptual models (LSP-W) described by
\cite{Krish2013Grounded}, as well as a fixed-structure neural module net (NMN). 
This improvement is particularly notable on the dataset with  quantifiers, where dynamic 
structure prediction produces a 20\% relative improvement  over the fixed baseline. 
A variety of predicted layouts are shown in \autoref{fig:geo:qualitative}.

\begin{figure}
\centering
\footnotesize
\tabulinesep=2mm
\begin{tabu}{|p{0.92\columnwidth}|}
\hline
Is Key Largo an island? \smallskip\newline
\mrl{(exists (and lookup[key-largo] find[island]))} \smallskip\newline
\correct{yes}: correct \\
\hline
What national parks are in Florida? \smallskip\newline
\mrl{(and find[park] (relate[in] lookup[florida]))} \smallskip\newline
\correct{everglades}: correct \\
\hline
What are some beaches in Florida? \smallskip\newline
\mrl{(exists (and lookup[beach] \newline
\strut~~~~~~~~~~~~~(relate[in] lookup[florida])))} \smallskip\newline
\incorrect{yes} (daytona-beach): wrong parse \\
\hline
What beach city is there in Florida? \smallskip\newline
\mrl{(and lookup[beach] lookup[city] \newline
\strut~~~~~(relate[in] lookup[florida]))} \smallskip\newline
\incorrect{[none]} (daytona-beach): wrong module behavior \\
\hline
\end{tabu}
\vspace{.5mm}
\caption{Example layouts and answers selected by the model on the GeoQA 
dataset. For incorrect predictions, the correct answer is shown in parentheses.}
\label{fig:geo:qualitative}
\vspace{-4mm}
\end{figure}

\section{Conclusion}
\label{sec:conclusion}

We have introduced a new model, the \emph{dynamic neural module network},
for answering queries about both structured and unstructured sources of
information. Given only (question, world, answer) triples as training data, 
the model learns to assemble neural networks on the fly from an inventory
of neural models, and simultaneously learns weights for these modules so
that they can be composed into novel structures. Our approach achieves
state-of-the-art results on two tasks.
We believe that the success of this work derives from two factors:

\emph{Continuous representations improve the expressiveness and learnability of
semantic parsers}: by replacing discrete predicates with differentiable neural
network fragments, we bypass the challenging combinatorial optimization problem
associated with induction of a semantic lexicon. In structured
world representations, neural predicate representations allow the model to
invent reusable attributes and relations not expressed in
the schema. Perhaps more importantly, we can extend
compositional question-answering machinery to complex, continuous world
representations like images.

\emph{Semantic structure prediction improves generalization in deep networks}:
by replacing a fixed network topology with a dynamic one, we can tailor the
computation performed to each problem instance, using deeper networks for more
complex questions and representing combinatorially many queries with
comparatively few parameters.  In practice, this results in considerable gains
in speed and sample efficiency, even with very little training data.

These observations are not limited to the question answering domain, and
we expect that they can be applied similarly to tasks like
instruction following, game playing, and language generation.

\section*{Acknowledgments}

JA is supported by a National Science Foundation Graduate Fellowship.
MR is supported by a fellowship within the FIT weltweit-Program
of the German Academic Exchange Service (DAAD). This
work was additionally supported by DARPA, AFRL, DoD
MURI award N000141110688, NSF awards IIS-1427425
and IIS-1212798, and the Berkeley Vision and Learning
Center.

\bibliography{jacob}

\begin{thebibliography}{}

\bibitem[\protect\citename{Andreas \bgroup et al.\egroup }2013]{Andreas13SPMT}
Jacob Andreas, Andreas Vlachos, and Stephen Clark.
\newblock 2013.
\newblock Semantic parsing as machine translation.
\newblock In {\em Proceedings of the Annual Meeting of the Association for
  Computational Linguistics}, Sofia, Bulgaria.

\bibitem[\protect\citename{Andreas \bgroup et al.\egroup }2016]{Andreas15NMN}
Jacob Andreas, Marcus Rohrbach, Trevor Darrell, and Dan Klein.
\newblock 2016.
\newblock Neural module networks.
\newblock In {\em Proceedings of the Conference on Computer Vision and Pattern
  Recognition}.

\bibitem[\protect\citename{Antol \bgroup et al.\egroup }2015]{Antol15VQA}
Stanislaw Antol, Aishwarya Agrawal, Jiasen Lu, Margaret Mitchell, Dhruv Batra,
  C~Lawrence Zitnick, and Devi Parikh.
\newblock 2015.
\newblock {VQA}: Visual question answering.
\newblock In {\em Proceedings of the International Conference on Computer
  Vision}.

\bibitem[\protect\citename{Beltagy \bgroup et al.\egroup
  }2013]{Beltagy13Markov}
Islam Beltagy, Cuong Chau, Gemma Boleda, Dan Garrette, Katrin Erk, and Raymond
  Mooney.
\newblock 2013.
\newblock Montague meets markov: Deep semantics with probabilistic logical
  form.
\newblock {\em Proceedings of the Joint Conference on Distributional and
  Logical Semantics}, pages 11--21.

\bibitem[\protect\citename{Berant and Liang}2014]{Berant14Paraphrasing}
Jonathan Berant and Percy Liang.
\newblock 2014.
\newblock Semantic parsing via paraphrasing.
\newblock In {\em Proceedings of the Annual Meeting of the Association for
  Computational Linguistics}, volume~7, page~92.

\bibitem[\protect\citename{Bordes \bgroup et al.\egroup
  }2014]{Bordes14GraphEmbedding}
Antoine Bordes, Sumit Chopra, and Jason Weston.
\newblock 2014.
\newblock Question answering with subgraph embeddings.
\newblock {\em Proceedings of the Conference on Empirical Methods in Natural
  Language Processing}.

\bibitem[\protect\citename{Bottou \bgroup et al.\egroup
  }1997]{Bottou97GraphTransformers}
L{\'e}on Bottou, Yoshua Bengio, and Yann Le~Cun.
\newblock 1997.
\newblock Global training of document processing systems using graph
  transformer networks.
\newblock In {\em Proceedings of the Conference on Computer Vision and Pattern
  Recognition}, pages 489--494. IEEE.

\bibitem[\protect\citename{Bottou}2014]{Bottou14Reasoning}
L{\'e}on Bottou.
\newblock 2014.
\newblock From machine learning to machine reasoning.
\newblock {\em Machine learning}, 94(2):133--149.

\bibitem[\protect\citename{De~Marneffe and Manning}2008]{DeMarneffe08Deps}
Marie-Catherine De~Marneffe and Christopher~D Manning.
\newblock 2008.
\newblock The {S}tanford typed dependencies representation.
\newblock In {\em Proceedings of the International Conference on Computational
  Linguistics}, pages 1--8.

\bibitem[\protect\citename{Grefenstette}2013]{Grefenstette13Logic}
Edward Grefenstette.
\newblock 2013.
\newblock Towards a formal distributional semantics: Simulating logical calculi
  with tensors.
\newblock {\em Joint Conference on Lexical and Computational Semantics}.

\bibitem[\protect\citename{Hermann \bgroup et al.\egroup }2015]{Hermann15AttQA}
Karl~Moritz Hermann, Tomas Kocisky, Edward Grefenstette, Lasse Espeholt, Will
  Kay, Mustafa Suleyman, and Phil Blunsom.
\newblock 2015.
\newblock Teaching machines to read and comprehend.
\newblock In {\em Advances in Neural Information Processing Systems}, pages
  1684--1692.

\bibitem[\protect\citename{Iyyer \bgroup et al.\egroup }2014]{Iyyer14Factoid}
Mohit Iyyer, Jordan Boyd-Graber, Leonardo Claudino, Richard Socher, and Hal
  {Daum\'e III}.
\newblock 2014.
\newblock A neural network for factoid question answering over paragraphs.
\newblock In {\em Proceedings of the Conference on Empirical Methods in Natural
  Language Processing}.

\bibitem[\protect\citename{Krishnamurthy and Kollar}2013]{Krish2013Grounded}
Jayant Krishnamurthy and Thomas Kollar.
\newblock 2013.
\newblock Jointly learning to parse and perceive: connecting natural language
  to the physical world.
\newblock {\em Transactions of the Association for Computational Linguistics}.

\bibitem[\protect\citename{Krishnamurthy and
  Mitchell}2013]{Krishnamurthy13CompVector}
Jayant Krishnamurthy and Tom Mitchell.
\newblock 2013.
\newblock Vector space semantic parsing: A framework for compositional vector
  space models.
\newblock In {\em Proceedings of the ACL Workshop on Continuous Vector Space
  Models and their Compositionality}.

\bibitem[\protect\citename{Kwiatkowski \bgroup et al.\egroup
  }2010]{Kwiatkowski10UBL}
Tom Kwiatkowski, Luke Zettlemoyer, Sharon Goldwater, and Mark Steedman.
\newblock 2010.
\newblock Inducing probabilistic {CCG} grammars from logical form with
  higher-order unification.
\newblock In {\em Proceedings of the Conference on Empirical Methods in Natural
  Language Processing}, pages 1223--1233, Cambridge, Massachusetts.

\bibitem[\protect\citename{Kwiatkowski \bgroup et al.\egroup
  }2013]{Kwiatkowski13Ontology}
Tom Kwiatkowski, Eunsol Choi, Yoav Artzi, and Luke Zettlemoyer.
\newblock 2013.
\newblock Scaling semantic parsers with on-the-fly ontology matching.
\newblock In {\em Proceedings of the Conference on Empirical Methods in Natural
  Language Processing}.

\bibitem[\protect\citename{Lewis and
  Steedman}2013]{Lewis13DistributionalLogical}
Mike Lewis and Mark Steedman.
\newblock 2013.
\newblock Combining distributional and logical semantics.
\newblock {\em Transactions of the Association for Computational Linguistics},
  1:179--192.

\bibitem[\protect\citename{Liang \bgroup et al.\egroup }2011]{Liang11DCS}
Percy Liang, Michael Jordan, and Dan Klein.
\newblock 2011.
\newblock Learning dependency-based compositional semantics.
\newblock In {\em Proceedings of the Human Language Technology Conference of
  the Association for Computational Linguistics}, pages 590--599, Portland,
  Oregon.

\bibitem[\protect\citename{Malinowski \bgroup et al.\egroup
  }2015]{Malinowski15VQA}
Mateusz Malinowski, Marcus Rohrbach, and Mario Fritz.
\newblock 2015.
\newblock Ask your neurons: A neural-based approach to answering questions
  about images.
\newblock In {\em Proceedings of the International Conference on Computer
  Vision}.

\bibitem[\protect\citename{Matuszek \bgroup et al.\egroup
  }2012]{Matuszek12Grounded}
Cynthia Matuszek, Nicholas FitzGerald, Luke Zettlemoyer, Liefeng Bo, and Dieter
  Fox.
\newblock 2012.
\newblock A joint model of language and perception for grounded attribute
  learning.
\newblock In {\em International Conference on Machine Learning}.

\bibitem[\protect\citename{Noh \bgroup et al.\egroup }2015]{Noh15DPPVQA}
Hyeonwoo Noh, Paul~Hongsuck Seo, and Bohyung Han.
\newblock 2015.
\newblock Image question answering using convolutional neural network with
  dynamic parameter prediction.
\newblock {\em arXiv preprint arXiv:1511.05756}.

\bibitem[\protect\citename{Pasupat and Liang}2015]{Pasupat15Tables}
Panupong Pasupat and Percy Liang.
\newblock 2015.
\newblock Compositional semantic parsing on semi-structured tables.
\newblock In {\em Proceedings of the Annual Meeting of the Association for
  Computational Linguistics}.

\bibitem[\protect\citename{Ren \bgroup et al.\egroup }2015]{Ren15VQA}
Mengye Ren, Ryan Kiros, and Richard Zemel.
\newblock 2015.
\newblock Exploring models and data for image question answering.
\newblock In {\em Advances in Neural Information Processing Systems}.

\bibitem[\protect\citename{Simonyan and Zisserman}2014]{Simonyan14VGG}
K~Simonyan and A~Zisserman.
\newblock 2014.
\newblock Very deep convolutional networks for large-scale image recognition.
\newblock {\em arXiv preprint arXiv:1409.1556}.

\bibitem[\protect\citename{Socher \bgroup et al.\egroup }2013]{Socher13CVG}
Richard Socher, John Bauer, Christopher~D. Manning, and Andrew~Y. Ng.
\newblock 2013.
\newblock Parsing with compositional vector grammars.
\newblock In {\em Proceedings of the Annual Meeting of the Association for
  Computational Linguistics}.

\bibitem[\protect\citename{Towell and Shavlik}1994]{Towell94KBNN}
Geoffrey~G Towell and Jude~W Shavlik.
\newblock 1994.
\newblock Knowledge-based artificial neural networks.
\newblock {\em Artificial Intelligence}, 70(1):119--165.

\bibitem[\protect\citename{Williams}1992]{Williams92Reinforce}
Ronald~J Williams.
\newblock 1992.
\newblock Simple statistical gradient-following algorithms for connectionist
  reinforcement learning.
\newblock {\em Machine learning}, 8(3-4):229--256.

\bibitem[\protect\citename{Wong and Mooney}2007]{Wong07WASP}
Yuk~Wah Wong and Raymond~J. Mooney.
\newblock 2007.
\newblock Learning synchronous grammars for semantic parsing with lambda
  calculus.
\newblock In {\em Proceedings of the Annual Meeting of the Association for
  Computational Linguistics}, volume~45, page 960.

\bibitem[\protect\citename{Xu and Saenko}2015]{Xu15AttVQA}
Huijuan Xu and Kate Saenko.
\newblock 2015.
\newblock Ask, attend and answer: Exploring question-guided spatial attention
  for visual question answering.
\newblock {\em arXiv preprint arXiv:1511.05234}.

\bibitem[\protect\citename{Xu \bgroup et al.\egroup }2015]{Xu15SAT}
Kelvin Xu, Jimmy Ba, Ryan Kiros, Kyunghyun Cho, Aaron Courville, Ruslan
  Salakhutdinov, Richard Zemel, and Yoshua Bengio.
\newblock 2015.
\newblock Show, attend and tell: Neural image caption generation with visual
  attention.
\newblock In {\em International Conference on Machine Learning}.

\bibitem[\protect\citename{Yang \bgroup et al.\egroup }2015]{Yang15AttVQA}
Zichao Yang, Xiaodong He, Jianfeng Gao, Li~Deng, and Alex Smola.
\newblock 2015.
\newblock Stacked attention networks for image question answering.
\newblock {\em arXiv preprint arXiv:1511.02274}.

\bibitem[\protect\citename{Yin \bgroup et al.\egroup }2015]{Yin15NeuralTable}
Pengcheng Yin, Zhengdong Lu, Hang Li, and Ben Kao.
\newblock 2015.
\newblock Neural enquirer: Learning to query tables.
\newblock {\em arXiv preprint arXiv:1512.00965}.

\bibitem[\protect\citename{Zeiler}2012]{Zeiler12Adadelta}
Matthew~D Zeiler.
\newblock 2012.
\newblock {ADADELTA}: {A}n adaptive learning rate method.
\newblock {\em arXiv preprint arXiv:1212.5701}.

\bibitem[\protect\citename{Zhou \bgroup et al.\egroup }2015]{Zhou15ClassVQA}
Bolei Zhou, Yuandong Tian, Sainbayar Sukhbaatar, Arthur Szlam, and Rob Fergus.
\newblock 2015.
\newblock Simple baseline for visual question answering.
\newblock {\em arXiv preprint arXiv:1512.02167}.

\end{thebibliography}
\bibliographystyle{naaclhlt2016}

\end{document}